\documentclass[journal]{IEEEtran} 
\usepackage{graphicx}
\usepackage{cite}
\usepackage{amsmath,amssymb,amsfonts}
\usepackage{algorithmic}
\usepackage{textcomp}
\usepackage{booktabs}
\usepackage{xspace}
\usepackage{svg}
\usepackage{amsmath}
\usepackage{footnote}
\usepackage{makecell}
\usepackage{multirow}
\usepackage{enumitem}
\usepackage{textcomp}
\def\BibTeX{{\rm B\kern-.05em{\sc i\kern-.025em b}\kern-.08em
    T\kern-.1667em\lower.7ex\hbox{E}\kern-.125emX}}

\usepackage{array}
\usepackage[nolist,nohyperlinks]{acronym}
\usepackage{hyperref}
\usepackage{subcaption}
\makesavenoteenv{tabular}

\usepackage[acronym]{glossaries}

\acrodef{GI}{gastrointestinal}
\acrodef{AI}{Artificial Intelligence} 
\acrodef{ML}{Machine Learning}
\acrodef{DL}{deep learning}
\acrodef{CNN}{Convolutional Neural Network}
\acrodef{CRC}{Colorectal Cancer}
\acrodef{WCE}{Wireless Capsule Endoscopy}  
\acrodef{FCN}{Fully Convolutional Network}
\acrodef{ASPP}{Atrous Spatial Pyramidal Pooling}
\acrodef{SGDR}{Stochastic Gradient Descent with Restart}
\acrodef{SGD}{Stochastic Gradient Descent}
\acrodef{MSE}{Mean Square Error}
\acrodef{ReLU}{Rectified Linear Unit}
\acrodef{BN}{Batch Normalization}
\acrodef{FPS}{Frame Per Second}
\acrodef{GAN}{Generative Adversarial Network}
\acrodef{mIoU}{mean Intersection over Union}
\acrodef{SOTA}{state-of-the-art}
\acrodef{DSC}{Dice Coefficient}
\acrodef{CADx}{Computer Aided Diagnosis}
\acrodef{ISIC}{international skin imaging collaboration}
\acrodef{RNN}{Recurrent  Neural  Network}
\acrodef{RL}{Recurrent Learning}
\acrodef{FCNN}{Fully Convolutional Neural Network}
\acrodef{TTA}{Test-Time Augmentation}
\acrodef{SE}{Squeeze and Excitation}
\acrodef{DRIVE}{Digital Retinal Images for Vessel Extraction}
\acrodef{FANet}{Feedback Attention Network}
\begin{document}
\title{{FANet:} A Feedback Attention Network for Improved Biomedical Image Segmentation}
\author{Nikhil Kumar Tomar, 
Debesh Jha, \IEEEmembership{Member, IEEE}, 
Michael A. Riegler, \IEEEmembership{Member, IEEE}, 
H{\aa}vard D. Johansen, \IEEEmembership{Member, IEEE},
Dag Johansen, \IEEEmembership{Member, IEEE}, 
Jens Rittscher, \IEEEmembership{Member, IEEE},
P{\aa}l Halvorsen, \IEEEmembership{Member, IEEE}, and 
Sharib Ali, \IEEEmembership{Member, IEEE}
%
\thanks{N. K. Tomar is with SimulaMet, Oslo, Norway}
\thanks{D. Jha is with SimulaMet, Oslo, Norway and UiT The Arctic University of Norway, Troms{\o}, Norway (corresponding email: debesh@simula.no).}
\thanks{M. A. Riegler is with SimulaMet, Oslo, Norway.}
\thanks{H.D. Johansen and D. Johansen are with UiT The Arctic University of Norway, Troms{\o}, Norway.}
\thanks{J. Rittscher is with the Department of Engineering Science, University of Oxford, Oxford, UK.}
\thanks{P. Halvorsen is with SimulaMet and Oslo Metropolitan University, Oslo, Norway.}
\thanks{S. Ali is with the Department of Engineering Science, University of Oxford, and Oxford NIHR Biomedical Research Centre, Oxford, UK (corresponding email: sharib.ali@eng.ox.ac.uk).}
}
\maketitle

\begin{abstract}
The increase of available large clinical and experimental datasets has contributed to a substantial amount of important contributions in the area of biomedical image analysis. Image segmentation, which is crucial for any quantitative analysis, has especially attracted attention. Recent hardware advancement has led to the success of deep learning approaches. However, although deep learning models are being trained on large datasets, existing methods do not use the information from different learning epochs effectively. In this work, we leverage the information of each training epoch to prune the prediction maps of the subsequent epochs. We propose a novel architecture called feedback attention network (FANet) that unifies the previous epoch mask with the feature map of the current training epoch. The previous epoch mask is then used to provide a hard attention to the learned feature maps at different convolutional layers. The network also allows to rectify the predictions in an iterative fashion during the test time. We show that our proposed \textit{feedback attention} model provides a substantial improvement on most segmentation metrics tested on seven publicly available biomedical imaging datasets demonstrating the effectiveness of FANet. The source code is available at \url{https://github.com/nikhilroxtomar/FANet}.

\end{abstract}


\IEEEpeerreviewmaketitle

\section{Introduction}
\label{sec:introduction}
\begin{figure*}[t!]
    \centering
    \includegraphics[width=\textwidth]{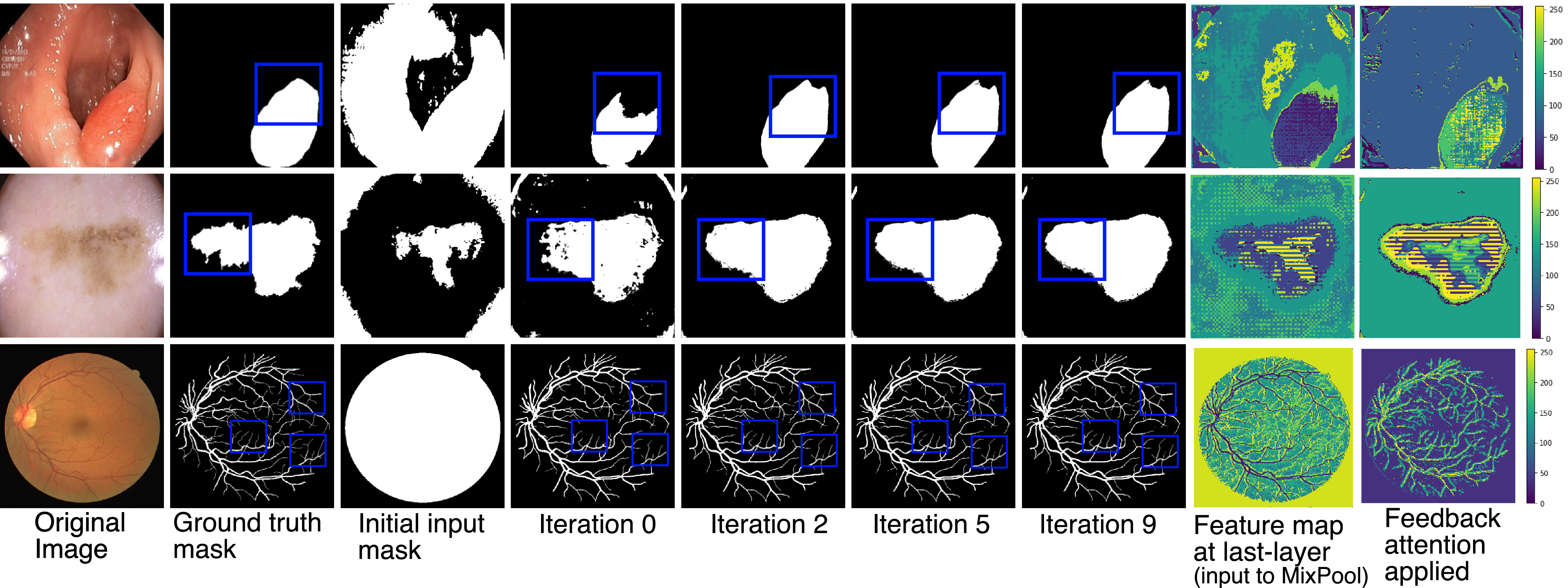}
    \caption{Semantic segmentation using our FANet architecture. Otsu thresholding is used for generating the initial mask used during $0^{th}$ iteration. Then the predictions are iteratively updated with the predicted mask. It can be observed that already at the 2$^{nd}$ iteration, the results converge. The corresponding feature maps before and after feedback attention at the last decoder layer of our FANet are shown as color images on the right.}\label{fig:fig0-examplary}
\end{figure*}

\IEEEPARstart{I}MAGE segmentation is one of the most studied problems in computer vision, where the main goal is to classify each pixel of an image to a specific class instance. This can either be pixels of any arbitrary objects such as cars or humans in natural scene data~\cite{chen2017deeplab}, satellite data in remote sensing~\cite{sun2021research, he2021dabnet}, or pixels of cancerous area or cells in biomedical imaging data~\cite{cardona2010integrated}. Substantial progress has been made in biomedical imaging due to which various modalities such as X-ray, Computerized Tomography (CT), Magnetic Resonance Imaging (MRI), endoscopy imaging, fundus imaging, Electron Microscopy (EM), and histology imaging exists.  While \ac{ML} methods usually provide improved performance over traditional computer vision methods, most of them require ground truth labels from domain experts, which are often scarce and may not represent enough variability in biomedical imaging data. This can affect \ac{ML} models resulting in only sub-optimal predictions. Furthermore, existing methods for semantic segmentation are based on a single-step prediction process that does not allow them to rectify their own predicted segmentation masks. Thus, these networks are constrained to only one set of learned weights that may not be enough to capture inter- and intra-class differences present in biomedical imaging data. In this work, we introduce an iterative approach that can refine the segmentation masks from previous mask predictions in a few iterative steps. This iteration process enables the network to steer towards the improved feature representation by taking advantage of subsequent attention mechanisms from previous mask, unlike classically used one-step segmentation methods~\cite{ronneberger2015u,chen2017deeplab}. Thus, aggregating these results over a few iterations provides improved segmentation masks (see illustration in Figure~\ref{fig:fig0-examplary}). 


Current developments of \acp{CNN}, \acp{RNN}, and attention modules have improved automated methods in biomedical image analysis. Widely used supervised end-to-end \acp{CNN} methods require a large and diverse training dataset to avoid overfitting. \acp{RNN} can be used to preserve the model compactness and can be effectively used for segmentation tasks in resource-constrained settings via an iterative update on internal states of network layers~\cite{Wang:ICCV2019}. However, they are known for their memory inefficient memory-bandwidth-bound computation and model complexity~\cite{bai2018empirical}. Additionally, spatial visual attention mechanisms used for image captioning for natural scene images~\cite{xu2015show} and for medical image segmentation~\cite{oktay2018attention} showed improvements in terms of both model convergence and performance metric. An attention mechanism allows networks to focus on a concrete class instance, thereby penalizing non-specific regions. Our model is thus inspired by the success of both visual attention mechanism and recurrent learning paradigm. 

{A} mask-guided contrastive attention model was used by Song et al.~\cite{Song_2018_CVPR} to deal with the background clutter. Unlike classical training mechanisms and motivated by the work of Song et al.~\cite{Song_2018_CVPR}, we propose to propagate the sample-specific mask output from the previous epoch to the successive epoch in a recursive fashion. Such a feedback mechanism can provide prior information that can help to learn {sample} variability, thereby enabling to train effectively on diverse datasets. Here, iterative prediction can be used to prune the predicted masks during the inference (see Figure~\ref{fig:fig0-examplary}). This allows {the network to learn both local and global features that can} rectify the mask output from the {learned} weights. Unlike \ac{TTA}~\cite{WANG201934}, where different transforms are utilized to mimic sample representations {and data diversity, we embed mask rectification during the training process}. To our knowledge, \ac{FANet} is the first deep learning model that {incorporates the ability to} self-rectify its predictions without requiring {heavy} transformations, ensemble strategies, and prior sample-specific knowledge. {FANet uses a} single end-to-end trainable network that allows information propagation during both train and test time.


{A} feedback mechanism {during the training is} central to our novel \ac{FANet} approach for semantic segmentation. The predicted map of each sample from the previous epoch {unified with the current state feature map} is used to provide attention. FANet uses an attention mechanism to different feature scales in the network, allowing it to capture variability in image samples. Additionally, our residual block with \ac{SE} layer allows us to improve channel interdependencies, which can be critical to tackle image quality issues. 
The main contributions of this work can be summarized as follows:
\begin{enumerate}
    
    \item \textbf{Feedback attention learning --- } A novel mechanism to utilize the variability present in each training sample. The mask outputs are propagated from one to subsequent epochs to supress the unwanted feature clutter.

   \item \textbf{Iterative refining of prediction masks --- } Using feedback information helps in refining the predicted masks in training as well as inference. During testing, we iterate over the input image and keep updating the input mask with the predicted mask {for up to 10 iterations (empirically set)}.
    \item \textbf{Embedded run-length encoding strategy --- } Binary mask outputs of each samples are efficiently compressed before being propagated to the next epoch. This provides a memory efficient mechanism for passing sample specific masks.
    \item \textbf{Systematic evaluation --- } Experiments on seven vastly different biomedical datasets suggest that FANet outperforms other \ac{SOTA} algorithms.

    \item \textbf{Efficient training --- } FANet achieves near \ac{SOTA} performance with far fewer training epochs. 
\end{enumerate}

\section{Related work}
\label{relatedwork}
In this section, we summarize relevant advances in medical image segmentation {and} feedback attention networks. {We also highlight} recent contributions to iterative refinement methods for image segmentation. 

\subsection{Biomedical image segmentation}
The basis of most modern \ac{CNN}-based semantic segmentation architectures are either \ac{FCN}~\cite{long2015fully} or an encoder-decoder architecture such as  U-Net~\cite{ronneberger2015u} originally designed for cell segmentation. Various modifications of these networks have been proposed both for semantic segmentation of natural images~\cite{zhao2017pyramid,wang2020deep} and biomedical image segmentation~\cite{zhou2018unet,oktay2018attention,zhou2019unet++,fan2020pranet,jha2019resunet++,jha2020doubleu,wang2020boundary}. In general, in the encoder, the image content is encoded using multiple convolutions to capture from low-level to high-level features, whereas in the decoder part of the network the prediction masks are obtained by multiple upsampling mechanism {or deconvolution operations}. Methods like PSPNet~\cite{zhao2017pyramid} and DeepLab~\cite{chen2017deeplab} incorporate convolutional feature maps of varying resolutions to segment both small and large-sized objects effectively. While PSPNet used a pyramid pooling module, DeepLab used \ac{ASPP} for encoding the multi-scale contextual information. Both PSPNet and DeepLab based architectures have been used widely in the medical imaging community {for biomedical image segmentation~\cite{hassan2020,sun2019colorectal}}.

\subsection{Feedback attention networks}
Visual attention has been widely used in computer vision for  pose estimation~\cite{Chu2017MulticontextAF}, {object detection~\cite{chen2017sca-cnn}, and image segmentation~\cite{CY2016Attention,ye2019cross}}. {Chu et al.~\cite{Chu2017MulticontextAF} incorporated the multi-context attention method into their end-to-end eight stack hourglass \ac{CNN} network where each sub-network of the hourglass generated a multi-resolution attention map}.  Attention mechanisms~\cite{LUNDERVOLD2019102,SCHLEMPER2019197} have also been utilized for posing explicit focus on the target region in medical imaging.  {Schlemper et al.~\cite{SCHLEMPER2019197} proposed a novel attention gate model that automatically learned to focus on the target structure of the varying shape and sizes by suppressing the irrelevant features and highlighting the silent feature for the specified medical image segmentation task}. Attention U-Net~\cite{oktay2018attention} used a gated operation in the U-Net architecture to focus on the target abdominal regions of CT datasets. Feedback mechanism for attention using two U-Net architectures with shared weights was used for cell segmentation~\cite{tsuda2020feedback,shibuya2020feedback}. The latter used a standard U-Net architecture with the second U-Net incorporating ConvLSTM~\cite{NIPS2015_07563a3f} to store the feature map (input-to-state) from the first U-Net network. However, feedback is only applied to the same epoch with state-to-state transitions. On the contrary, our approach utilizes a feedback mechanism that propagates information flow from the previous epoch to the current epoch in an attention mechanism. We employ the predicted masks from the previous epoch as hard attention to prune the segmentation output.
\subsection{Iterative refinement for segmentation}
An iterative refinement of the segmentation mask by feeding the input image and the predicted segmentation mask to a modified U-Net architecture was done by Mosinska et al.~\cite{mosinska2018beyond}. {In this work, the authors used an iterative refinement pipeline to enhance the quality of the predicted segmentation mask}. Similarly, iterative update of latent space and minimization of the Structure Similarity Index Measure (SSIM) loss was used to refine the predicted segmentation maps during test time in~\cite{Prashant:ECCV20}. Recently, iterative refinement strategies have also been used for pose estimation~\cite{newell2016stacked,wei2016convolutional} that used consecutive modules for refinement of the predictions with a loss function for the evaluation of output in each module. These iterative refinement processes show improved predictions and are able to handle domain shifts or object shape variability without requiring very deep networks~\cite{Prashant:ECCV20}. However, a major bottleneck in these methods is the requirement of a large number of iterations for model convergence. Unlike these methods, our proposed \ac{FANet} provides attention to the specific region-of-interest and can prune the {predicted segmentation masks} in less than ten iterations without requiring any optimization scheme. 

\section{Method}
\label{methodology}
\begin{figure*}[t!]
    \centering
    \includegraphics[width=0.9\textwidth]{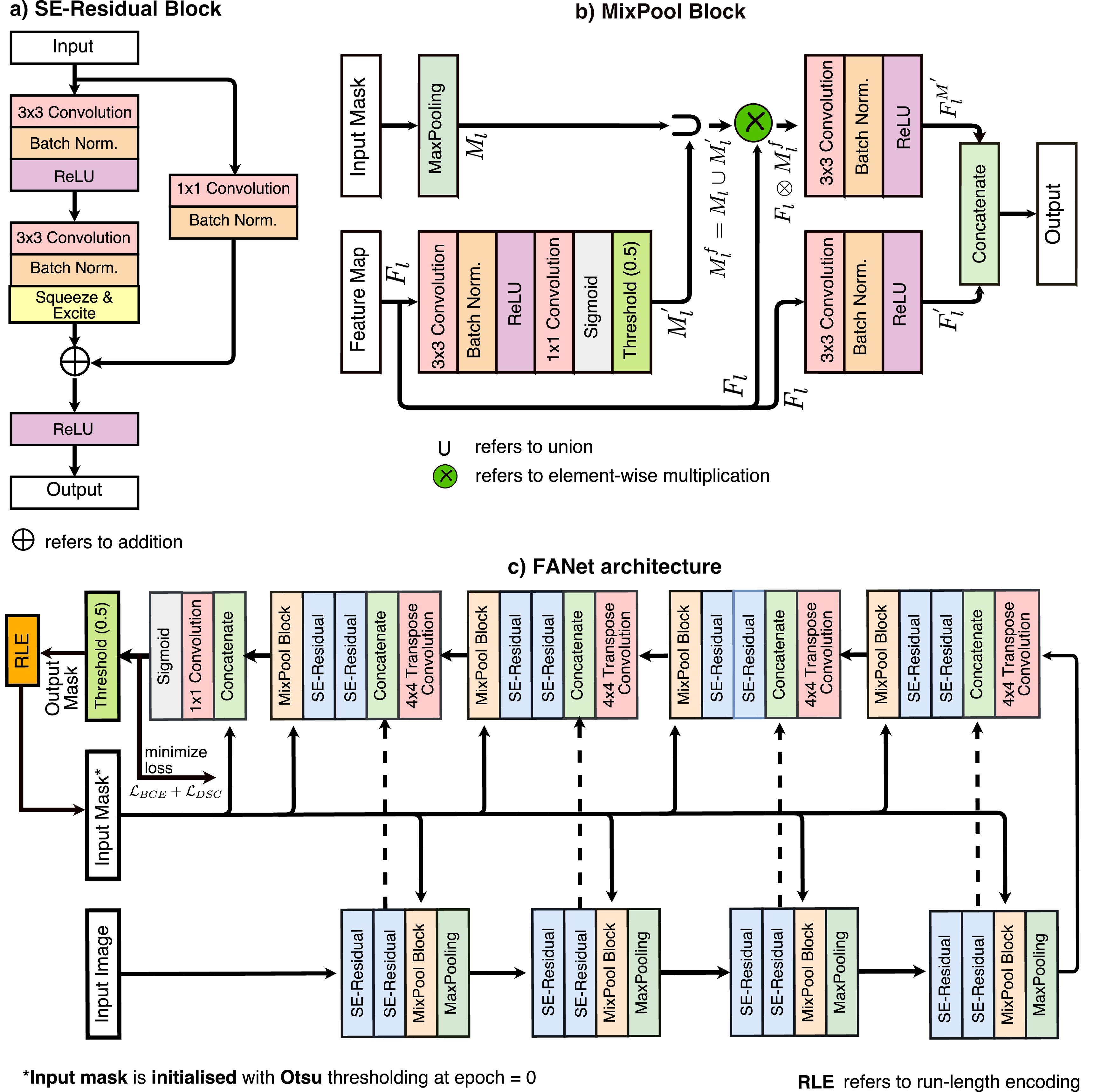}
    \caption{FANet with squeeze and excite residual (SE-Residual) block and MixPool block. (a) SE-Residual block integrated with squeeze and excite layer uses $1\times 1$ convolution to concatenate the high-resolution feature representation with the the encoded feature vector. (b) MixPool block represents attention mechanism in our network. The input mask is downscaled to the corresponding layer feature map size $M_l$ which is fused with the masked feature map representation $M_l^{'}$ for hard attention of input feature in that layer $F_l$. Finally, the attenuated feature map $F_l^{M^{'}}$ and the feature maps $F_l^{{'}}$ are both concatenated. c) Proposed FANet showing the complete network architecture. Encoder-decoder architecture with skip-connections (in dotted arrows) from SE-Residual blocks to preserve high- and intermediate resolution feature representations and MixPool block connections (with solid arrows) that allow to feedback the previous mask predictions.} 
    \label{fig:proposedarchitecture}
\end{figure*}

{In this section, we describe the components of the proposed \ac{FANet} architecture.} The overall design along with the proposed feedback attention learning mechanism is illustrated in  Figure~\ref{fig:proposedarchitecture}. 

\begin{table*} [t!]
\footnotesize
 \caption{Details of the biomedical datasets used in our experiments. ``Train", ``Train after aug." and ``Test" denote the number of training samples, number of training samples after image augmentation, and number of test samples, respectively.}
    \label{table:datasettable}
    \centering
          \begin{tabular}{@{}l|l|l|l|l|l|l@{}} 
            \toprule
        \textbf{Dataset} &\textbf{Images} &\textbf{Size}  &\textbf{Train} &\textbf{Train after aug.} & \textbf{Test} &\textbf{Application}\\ 
\midrule
Kvasir-SEG~\cite{jha2020kvasir} & 1000 & Variable & 880 & 16720 & 120 &Colonoscopy\\  
CVC-ClinicDB~\cite{bernal2015wm} & 612 & $384 \times 288$ & 490 & 14210 & 61 &Colonoscopy\\  
2018 Data Science Bowl~\cite{caicedo2019nucleus} & 670 & $256\times 256$ & 335 & 10720 & 134 & Nuclie \\  
ISIC 2018 (Lesion Boundary Segmentation)  ~\cite{codella2018skin,tschandl2018ham10000}& 2596 & Variable & 1815 & 39930 & 259 & Dermoscopy\\ 
EM dataset~\cite{cardona2010integrated} & 30 & $512\times 512$ & 24 & 384 & 3 & Cell \\
DRIVE Database~\cite{staal2004ridge} & 40 & $584 \times 565$ & 20 & 640 & 20 & Retina Vessel\\
CHASE-DB1~\cite{owen2009measuring} & 28 & $584 \times 565$ & 20 & 640 & 8 & Retina Vessel\\
        \bottomrule
\end{tabular}
\end{table*}	

\subsection{SE-Residual block}
Deeper networks improve the performance of the model significantly, but an increase in depth can cause either vanishing or exploding gradients problem~\cite{he2016deep}. To deal with this, we take advantage of shortcut connections between layers in the residual learning paradigm. Our SE-Residual block uses two 3$\times$3 convolutions and an identity mapping, where each convolution layer is followed by a \ac{BN} layer and a \ac{ReLU} non-linear activation function. The identity mapping is used to connect the input and the output of the convolution layer (Figure~\ref{fig:proposedarchitecture} a). 

Similar to the work by Hu et al.~\cite{Hu_2018_CVPR}, we add a \ac{SE} layer in the residual network. The \ac{SE} layer acts as a content-aware mechanism that re-weights each channel accordingly to create robust representations. Hence, it allows the network to become more sensitive to significant features while suppressing irrelevant features. This goal is accomplished in two steps.~First, the feature maps are squeezed by using the global average pooling to get a global understanding of each channel. The squeeze operation results in a  feature vector of size $\mathit{n}$, where $\mathit{n}$ refers to the number of channels. In the second step: excitation, this feature vector is feed through a two-layered feed-forward neural network, where the number of features is first reduced and then expanded to the original size $\mathit{n}$. Now, this $\mathit{n}$ sized vector represents the weight of the original feature maps, which is used to scale each channel.

\subsection{MixPool block}
The proposed MixPool block shown in Figure~\ref{fig:proposedarchitecture} (b) is used in multiple layers of our FANet architecture. This block facilitates the flow of sample-wise feedback information between consecutive epochs providing a hard attention to the learned features from SE-Residual block. The layer provides focus to the relevant features in both contraction path and expansion path layers.  The ‘hard’ attention map consists of the values 0 and 1, i.e., attention to a specific region only unlike soft attention where the probability map is estimated. The advantage of hard attention it that it allows to keep only the important features and ignore irrelevant features. During the element-wise multiplication, the values from the input feature map, if multiplied by 0, becomes 0, leaving the essential features for further operations. Another advantage of such methods is their computational speed, scalability, and ease of interpretation~\cite{xu2015show,malinowski2018learning}. The input mask used during training is compressed using the run-length encoding technique to save the memory footprint. 

As in Figure~\ref{fig:proposedarchitecture} (b-c), fist feature maps from the SE-Residual blocks $F_l$ in each layer is passed through a $3 \times 3$ convolution followed by a \ac{BN} and a \ac{ReLU} activation function. Then, we apply a $1 \times 1$ convolution and a sigmoid activation function $\sigma(\cdot)$ with a threshold of 0.5 to obtain the binary mask $M_{l}^{'}$ to contribute to the \textit{spatial attention map generation} given by:

\begin{equation}
      M_{l}^{'} = \sigma\left(\mathit{conv}(F_{l})\right)= \begin{cases}
    1,& \text{if } \sigma(\cdot)\geq 0.5\\
    0,              & \text{otherwise}. 
    \end{cases} \label{eq:1}\\
\end{equation}
Secondly, we apply appropriate max-pooling on the input mask (from the previous epoch) and resize it to the size of the spatial attention map  $M_{l}^{'}$. A union operation is then applied between the resized mask and the spatial attention map. This confirms that we obtain the feature from both the feedback and the spatial attention maps to further create a new unified spatial attention map.
Next, an element-wise multiplication operation is applied between the unified mask and the original feature map that suppresses the irrelevant features and enhances the important ones. The enhanced and the original feature maps are then followed by a $3 \times 3$ convolution, \ac{BN}, and a \ac{ReLU}. These operations are used to improve the network’s ability to learn non-linearity in the model prediction.

Finally, we concatenate the output of both activation functions, which constitutes the output of our MixPool block given by:
\begin{align}
     \text{Output}_{\text{MixPool}} = F_{l}^{'\frown} \left(F_l\otimes(M_l\cup M_{l}^{'})\right)^{'},\label{eq:2}
\end{align}
\noindent{where} $\frown$ denotes the concatenation operator, $\otimes$ is element-wise multiplication, and $\cup$ represents the union operation.

\subsection{Proposed FANet architecture}
\label{Proposedarchitecture}
The block diagram of FANet is illustrated in Figure~\ref{fig:proposedarchitecture} (c). It uses an encoder-decoder design common to many semantic segmentation architectures. We combine the strength of a residual network enhanced with \ac{SE} as SE-Residual block, and MixPool block that facilitates the attention and propagation of information flow from the current learning paradigm and that of the previous epoch. We implement a recurrent learning mechanism in both encoder and decoder layers that allows to achieve efficient segmentation. The MixPool block uses the previous segmentation map (as an input mask through RLE encoding), which contains the information from prior training and uses it to improve the semantic representation of the feature maps.

We first use the Otsu thresholding~\cite{otsu1979threshold} to generate an initial input mask for training the proposed architectural model. The variability in the input mask is refined over the training epochs and the model learns over time to prune input or previous epoch masks with learned semantically meaningful features together. To achieve this, we use the novel MixPool block that uses the input mask and applies hard attention over the subsequent input feature maps. The hard attention enables the network to highlight semantically meaningful features for the target region-of-interest in the entire network. The network thus not only learns to predict features maps but also strengthens a joint pruning mechanism that is dependent on the input mask. As a result, the devised network is able to rectify the predicted segmentation maps in an iterative fashion unlike conventional methods which do not have such pruning ability. This provides a strong rational behind our work that is applicable beyond single step inference prediction with capability of refining prediction maps.

The proposed network architecture is a \ac{FCNN} consisting of four encoder and four decoder blocks. The encoder takes the input image, downsamples it gradually, and encodes it in a compact representation. Then, the decoder takes this compact representation and tries to reconstruct the semantic representation by gradually upsampling it and combining the features from the encoder. Finally,  we receive a pixel-wise categorization of the input image. Both the encoder and the decoder are built using the SE-Residual block, and an additional concatenation of the original resolution feature representation in the encoder is added at each resolution scale. This mechanism minimizes the loss of feature representations during downscaling and upscaling processes. 

Each encoder network starts with two SE-Residual blocks, which consist of two $3\times3$ convolutions and a shortcut connection known as identity mapping, connecting the input and output of the two convolution layers. Each convolution is followed by a \ac{BN} and a \ac{ReLU} activation function. The output of the second SE-Residual block acts as skip connection for the corresponding decoder block. After that, it is followed by the MixPool block, which has the previous epoch segmentation mask and provides a hard-attention over the incoming feature maps. This process is repeated for each of the downscaled layers. 

Each decoder network starts with a $4\times 4$ transpose convolution that doubles the spatial dimensions of the incoming feature maps. These feature maps are concatenated with feature maps from the corresponding encoder block through skip connections. The skip connections help to propagate the information from the upper layers, which are sometimes lost due to the depth of the network. The skip connections are followed by two SE-Residual blocks, which help to eliminate the problem of vanishing gradient. The MixPool block that utilizes the segmentation mask from the previous epoch is then applied creating a hard-attention over the learned feature maps. Next, we concatenate the feature maps from the last decoder block and the segmentation mask from the previous epoch. Finally, we apply a $1\times 1$ convolution with the sigmoid activation function. The output of this is used to both minimize the training loss, using a combined binary cross-entropy and dice loss, and to generate segmentation masks that are stored as a run-length encoded compression for each sample and propagated during the next epoch. The RLE is updated after each epoch. 
%
Similarly, the network learns to adapt the weights in iterative training, this mechanism is also utilized during the test time. As shown in Figure~\ref{fig:fig0-examplary}, test results are pruned in a few iterations during the test time. Unlike many methods in literature~\cite{mosinska2018beyond,Prashant:ECCV20}, we utilize the same network without any complementary loss function optimization. 

\section{Experiments}
\label{experiments}
\subsection{Setup}
\subsubsection{Dataset and Evaluation Metrics}
To evaluate the proposed architecture, we have selected seven datasets that capture different segmentation tasks in biomedical imaging. The details of each dataset can be found in Table~\ref{table:datasettable}. The dataset images contain the images of organs and lesions acquired under different imaging protocols. For the retina vessel segmentation task, we use DRIVE and CHASE-DB1 datasets. These two datasets are aimed at various diseases related to diseases of retina vessels, such as  retinopathy, retinal vein occlusion, and retinal artery occlusion. The ISIC 2018 dataset, which is a dermoscopy dataset that is useful in the diagnosis of skin cancer, is the third dataset focused on medical imaging data. This dataset contains a wide variety of skin cancer images of different sizes and shapes, which helps in a better understanding of the disease. We have further included Kvasir-SEG and CVC-ClinicDB colonoscopy datasets. These datasets contain the image frames extracted from different colonoscopy interventions and are focused on colorectal polyps that are one of the cancer precursors in the colon and rectum. It highly increases the chance of avoiding lethal cancer by early detection. In addition, we have included two datasets acquired from biological imaging aimed at understanding of the cellular processes. These include the 2018 Data Science Bowl and the EM datasets. The 2018 Data Science Bowl dataset contains images with a large number of variable shaped nuclei acquired from different cell types, magnification, and imaging modalities. This dataset is designed for automated nuclei segmentation. Similarly, the EM dataset contains the transmission EM images of the neural structures of the Drosophila nerve cord. This dataset is aimed at the automated segmentation of the neural structures. All experiments on these datasets are conducted on the same train, validation, and test splits as provided by the previously published works reported in this paper.

To evaluate \ac{SOTA} deep learning methods and our proposed FANet, we have used standard evaluation metrics that includes \ac{DSC} (a.k.a. F1), \ac{mIoU}, precision, and recall. We have additionally calculated specificity for those datasets where this metric was previously used for benchmarking.  

\subsubsection{Implementation details}
All the training is performed on a Volta 100 GPU and an NVIDIA DGX-2 system using the PyTorch 1.6. framework. For test inference, we have used an NVIDIA GTX 1050 Ti GPU for our method and all \ac{SOTA} methods used in the paper as this hardware is widely available. Our model is trained for 100 epochs \textit{(empirically set)} using an Adam optimizer with a learning rate of $1e^{-4}$ for all the experiments except for the \ac{DRIVE} and the CHASE-DB1 dataset where the learning rate was adjusted to $1e^{-3}$ due to the small size of the training dataset. Datasets were chosen such that the efficiency of our model could be compared to the \ac{SOTA} methods. A combination of binary cross-entropy and dice loss has been used as the loss function. ReduceLROnPlateau callback was used to monitor the learning rate and adjust it to obtain optimal training performance. All the images used in the study were resized to $512\times 512$ except for the 2018 Data Science Bowl and the CVC-ClinicDB dataset, where images were resized to $256\times256$. Data augmentation, such as random crop, flipping, rotation, elastic transformation, grid distortion, optical distortion, grayscale conversion, random brightness, contrast, channel, and course dropout were used. 

\subsubsection{Ablation study}
In order to evaluate the strength of our proposed FANet architecture, we perform a thorough ablation study. For this, we have used all seven datasets and evaluated on several metrics for baseline (FANet without MixPool), baseline with MixPool, and the combination of baseline, MixPool, and feedback (proposed).
\subsection{Results}
\label{result}
Below we present quantitative results on seven different biomedical imaging datasets and compare with corresponding \ac{SOTA} methods. 
\subsubsection{Results on Kvasir-SEG} 
\begin{table}[!t]
\footnotesize
\caption{Results on the Kvasir-SEG~\cite{jha2020kvasir}.}
\begin{tabular}{@{}l|l|l|l|l|l@{}}
\toprule
\textbf{Method} &\textbf{Backbone} &\textbf{F1} &\textbf{mIoU} &\textbf{Recall}& \textbf{Prec.} \\ 
\midrule
{U-Net}~\cite{ronneberger2015u} & - & 0.5969 & 0.4713 & 0.6171 & 0.6722 \\ 
ResUNet~\cite{zhang2018road} & - & 0.6902 & 0.5721 & 0.7248 & 0.7454 \\ 
{ResUNet++}~\cite{jha2019resunet++} & - & 0.7143 & 0.6126 & 0.7419 & 0.7836 \\
{FCN8}~\cite{long2015fully}  &  VGG 16 & 0.8310 & 0.7365 & 0.8346 & 0.8817 \\
HRNet~\cite{wang2020deep}& - & 0.8446 & 0.7592 & 0.8588 & 0.8778 \\ 
{DoubleU-Net}~\cite{jha2020doubleu} & VGG 19 & 0.8129 & 0.7332 & 0.8402 & 0.8611 \\ 
{PSPNet}~\cite{zhao2017pyramid}  & ResNet50 & 0.8406 & 0.7444 & 0.8357 & 0.8901 \\ 
DeepLabv3+~\cite{chen2018encoder} & MobileNet & 0.8425 & 0.7575 & 0.8377 & 0.9014 \\ 
{DeepLabv3+}~\cite{chen2018encoder} & ResNet50 & 0.8572 & 0.7759 & 0.8616 & 0.8907 \\ 
{DeepLabv3+}~\cite{chen2018encoder}& ResNet101 & 0.8643 & 0.7862 & 0.8592 & \textbf{0.9064} \\ 
U-Net~\cite{ronneberger2015u} & VGG19 & 0.7535 & 0.6571 & 0.7364 & 0.8565 \\ 
U-Net++~\cite{zhou2018unet++} & - & 0.8002 & 0.7000 & 0.8716 & 0.7992 \\ 
Attention U-Net~\cite{oktay2018attention} & - & 0.7944 & 0.6959 & 0.8383 & 0.8287 \\ 
\textbf{FANet} & - & \textbf{0.8803} & \textbf{0.8153} & \textbf{0.9058} & 0.9005 \\ 
\bottomrule
\end{tabular}
\label{tab:kvasirseg}
\end{table}

Kvasir-SEG~\cite{jha2020kvasir} is a publicly available polyp segmentation dataset acquired from clinical colonoscopy procedures. This dataset has been widely used for algorithm benchmarking. We have trained our model and compared it with recent \ac{SOTA} methods on Kvasir-SEG. A comparison with widely accepted segmentation methods with different backbones (see Table~\ref{tab:kvasirseg}) shows that our approach is improved performance compared to the \ac{SOTA} methods (on the same train-test split). Our FANet outperforms all the \ac{SOTA} methods on almost all metrics. While outperforming most U-Net and its variants, it can be observed that FANet achieved an F1 score of 0.8803, which is 1.6\% and 3.57\% better than the most accurate DeepLabv3+ with ResNet101 backbone and the recent HRNet. 
%
\subsubsection{Results on CVC-ClinicDB dataset}
\begin{table}[!t]
\footnotesize
\centering
\caption{Results on the CVC-ClinicDB~\cite{bernal2015wm}.}
\begin{tabular}{@{}l|l|l|l|l@{}}
\toprule
\textbf{Method} & \textbf{F1} & \textbf{mIoU }& \textbf{Recall} & \textbf{Precision} \\ 
\midrule
U-Net (MICCAI’15)~\cite{ronneberger2015u} & 0.8230 & 0.7550 & -  & - \\ 
ResUNet-mod~\cite{zhang2018road} & 0.7788 & 0.4545 & 0.6683 & 0.8877 \\ 
ResUNet++~\cite{jha2019resunet++}  & 0.7955 & 0.7962 & 0.7022 & 0.8785 \\ 
SFA~(MICCAI’19)~\cite{fang2019selective} & 0.7000 & 0.6070 & - & - \\ 
PraNet~\cite{fan2020pranet} & 0.8990 & 0.8490 & - & - \\ 
U-Net++~\cite{zhou2018unet++} &0.9377 &0.8890 &0.9405 &0.9432  \\ 
Attention U-Net~\cite{oktay2018attention} &0.9325 &0.8856 &0.9276 &0.9546  \\ 
\textbf{FANet} &\textbf{0.9355} & \textbf{0.8937} & \textbf{0.9339} & \textbf{0.9401} \\ 
\bottomrule
\end{tabular}
\label{tab:cvclinicDB}

\end{table}
CVC-ClinicDB is another commonly used dataset for colonoscopy image analysis. FANet architecture outperforms all the \ac{SOTA} methods on this dataset by a large margin with F1 of 0.9355, \ac{mIoU} of 0.8937, recall of 0.9339, and precision of 0.9401 (see Table~\ref{tab:cvclinicDB}). FANet achieves the best trade-off between recall and precision compared to the ResUNet-based architectures~\cite{zhang2018road,jha2019resunet++}. The strength of the FANet can be observed by the large improvement of 23.17\% in the recall and 5.24\% in the precision over the \ac{SOTA} ResUNet++~\cite{jha2019resunet++}. The recall suggests that our method is more clinically preferable than the \ac{SOTA}. A higher recall is desired in the systems used for clinical diagnosis~\cite{gilvary2019missing}.
\subsubsection{Results on 2018 Data Science Bowl} 
\begin{table}[!t]
\footnotesize
\centering
\caption{Results on the 2018 Data Science Bowl~\cite{caicedo2019nucleus}.}
\begin{tabular}{@{}l|l|l|l|l|l@{}}
\toprule
\textbf{Method} &\textbf{Backbone} & \textbf{F1}  & \textbf{mIoU} &\textbf{Recall}& \textbf{Prec.} \\
\midrule
U-Net~\cite{ronneberger2015u} & ResNet101  & 0.7573 & 0.9103 & - & -  \\ 
DoubleU-Net~\cite{jha2020doubleu} & VGG19  & 0.7683 & 0.8407 & 0.6407 & \textbf{0.9596}  \\
U-Net++~\cite{zhou2018unet++} & - &0.9117 &0.8477 &0.9203 &0.9107  \\ 
Attention U-Net~\cite{oktay2018attention} & - &0.9179 &0.8570 &0.9183 &0.9235  \\ 
\textbf{FANet} & None  &\textbf{0.9176}& 0.8569 & \textbf{0.9222} & 0.9194\\ 
\bottomrule
\end{tabular}
\label{tab:DSB}
\end{table}
Cell nuclei segmentation in microscopy imaging is a common task in the biological image analysis~\cite{caicedo2019nucleus}. We used the publicly available 2018 Data Science Bowl (DSB) challenge dataset and compared our results with the \ac{SOTA} methods. Table~\ref{tab:DSB} shows that FANet produces an F1 of 0.9176, \ac{mIoU} of 0.8569, and recall of 0.9222 with an improvement of 2.02\% in F1 with respect to \ac{SOTA} UNet++~\cite{zhou2019unet++} and 28.15\% improvement in recall compared to the best performing DoubleU-Net~\cite{jha2020doubleu}. In general, FANet achieves the best trade-off between precision and recall compared to the \ac{SOTA} methods resulting in the highest F1 score (0.9176). The qualitative results with 2018 DSB also show that the predicted FANet produces high-quality segmentation masks for cell nuclei with respect to the ground truth (see Figure~\ref{fig:qualitativeresult}). 
%

\subsubsection{Results on ISIC 2018 dataset}
\begin{table}[!t]
\footnotesize
\centering
\caption{Results on the ISIC 2018 (Skin Caner Segmentation)~\cite{codella2018skin,tschandl2018ham10000}.}
\begin{tabular}{@{}l|l|l|l|l|l@{}}
\toprule
\textbf{Method} & \textbf{F1} & \textbf{mIoU }& \textbf{Recall}& \textbf{Spec.} & \textbf{Prec.} \\ 
\midrule
U-Net~\cite{ronneberger2015u} & 0.6740  & 0.5490 & 0.7080 & 0.9640 & -    \\ 
R2U-Net~\cite{alom2018recurrent} & 0.6790  & 0.5810 & 0.7920 & 0.9280 & -    \\ 
Attention R2U-Net~\cite{alom2018recurrent} &  0.6910  & 0.5920 & 0.7260 & 0.9710 & -    \\ 
BCDU-Net (d=1)~\cite{azad2019bi} & 0.8470  & - & 0.7830 & 0.9800 & -    \\ 
BCDU-Net (d=3)~\cite{azad2019bi} & 0.8510  & -  & 0.7850 & \textbf{0.9820} & -       \\ 
U-Net++~\cite{zhou2018unet++} &0.8088 &0.7319 &0.8450 &0.9110 &0.8648 \\ 
Attention U-Net~\cite{oktay2018attention} & 0.8205 &0.7346 &0.8516 &0.9135 &0.8645  \\ 
\textbf{FANet} & \textbf{0.8731} & \textbf{0.8023} & \textbf{0.8650} & 0.9611 & \textbf{0.9235} \\ 
\bottomrule
\end{tabular}
\label{tab:ISIC2018}
\end{table}

Skin cancer is one of the most commonly diagnosed cancers in the US. Early detection of melanoma can improve the five-year survival rate and help prevent it in 99\% of the cases~\cite{cancerfigures}. Table~\ref{tab:ISIC2018} shows the results on the publicly available International Skin Imaging Collaboration (ISIC) 2018 dataset. FANet outperformed all the methods on almost all evaluation metrics (F1, mIoU, and recall). FANet achieved 0.8731 on F1 and recall of 0.8650 with an improvement of 2.21\% and 8.00\%, respectively, over the most accurate \ac{SOTA} BCDU-Net (d=3) method. A competitive specificity and precision were also recorded. From the qualitative results in Figure~\ref{fig:qualitativeresult}, we can see that the input mask produced by Otsu thresholding shows under segmentation, which is improved significantly using FANet. The masks produced by FANet have smooth boundaries. 

\subsubsection{Results on DRIVE dataset}
\begin{table}[!t]
\footnotesize
\centering
\caption{Results on the DRIVE dataset~\cite{staal2004ridge}.}
\begin{tabular}{@{}l|l|l|l|l|l@{}}
\toprule
\textbf{Method} & \textbf{F1} & \textbf{mIoU} & \textbf{Recall}& \textbf{Spec.} & \textbf{Prec.} \\ 
\midrule
U-Net~\cite{ronneberger2015u} & 0.8174 & - & 0.7822 & 0.9808 & - \\ 
Residual U-Net~\cite{alom2018recurrent} & 0.8149 & - & 0.7726 & 0.9820 & - \\ 
Recurrent U-Net~\cite{alom2018recurrent}  & 0.8155 & - & 0.7751 & 0.9816 & - \\ 
R2U-Net~\cite{alom2018recurrent} &0.8171 & - &0.7792 &0.9813 & - \\ 
DenseBlock-UNet~\cite{yao2020eye3dvas} &0.8146 & - &0.7928 &0.9776 & - \\ 
DUNet~\cite{jin2019dunet} & 0.8190 & - &0.7863 &0.9805 & - \\ 
IterNet~\cite{li2020iternet} &\textbf{0.8218} &- &0.7791 &0.9831 & -  \\ 
IterNet(Patched)~\cite{li2020iternet} &0.8205 &-  &0.7235 & \textbf{0.9838} &- \\ 
U-Net++~\cite{zhou2018unet++} &0.7960 &0.6615 &0.7903 &0.9818 &0.8070 \\ 
Attention U-Net~\cite{oktay2018attention} &0.7984 &0.6648 &0.7877 &0.9827 &0.8146 \\ 
\textbf{FANet} &0.8183 & \textbf{0.6927} & \textbf{0.8215} &0.9826 & \textbf{0.8189} \\ 
\bottomrule
\end{tabular}
\label{tab:Drivedataset}
\end{table}

\begin{figure}[t!]
    \centering
    \includegraphics[clip, width = \columnwidth]{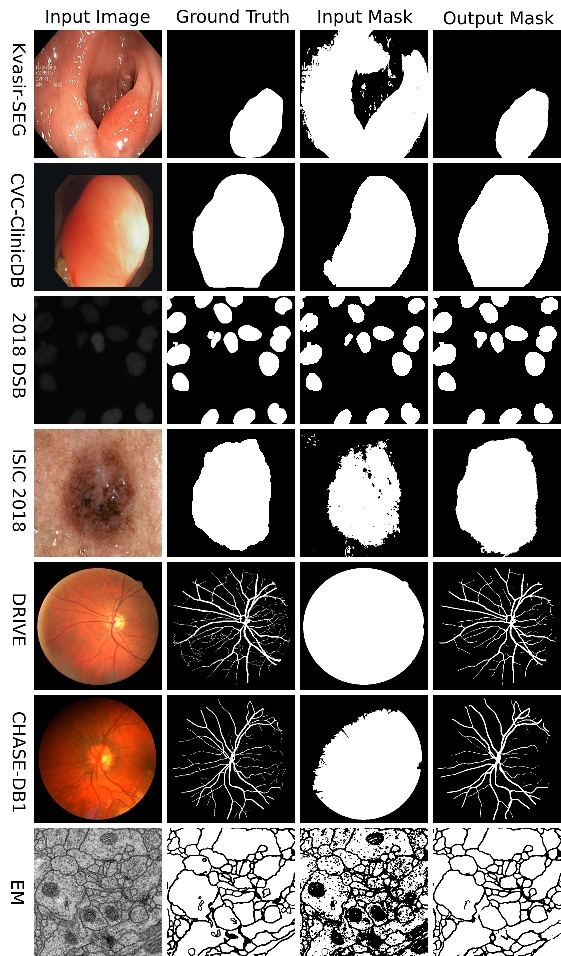}
    \caption{Qualitative results of FANet on seven biomedical image segmentation datasets. {The initial ``input mask'' is generated using \textit{Otsu thresholding}}. The ``output mask'' is the predicted segmentation mask from the  FANet model.}
    \label{fig:qualitativeresult}
\end{figure}

The automated segmentation of vessels in fundus images can assist in the diagnosis and treatment of diabetic retinopathy. The quantitative result on the publicly available DRIVE dataset is presented in Table~\ref{tab:Drivedataset}. We can observe that the proposed FANet achieves an F1 score of 0.8183, \ac{mIoU} of 0.6927, recall of 0.8215, and precision of 0.8189. The proposed method achieves an improvement of 4.24\% in the recall over \ac{SOTA} IterNet~\cite{li2020iternet}. Although the F1 of the IterNet is 0.35\% higher than FANet, the recall is relatively lower, and other metrics such as \ac{mIoU} and precision are not presented. For our proposed FANet, the precision of 0.8189 is well balanced with the obtained recall. The higher recall produced by FANet shows that our method is more clinically relevant. The quality of the segmentation masks in Figure\ref{fig:qualitativeresult} demonstrates the efficiency of FANet.

\subsubsection{Results on CHASE-DB1 dataset}
\begin{table}[!t]
\footnotesize
\centering
\caption{Results on the CHASE-DB1 dataset~\cite{owen2009measuring}.}
\begin{tabular}{@{}l|l|l|l|l|l@{}}
\toprule
\textbf{Method} & \textbf{F1} & \textbf{mIoU }& \textbf{Recall}& \textbf{Spec.} & \textbf{Prec.} \\ 
\midrule
U-Net~\cite{ronneberger2015u} & 0.7993 & - & 0.7840 & 0.9880 & - \\ 
DenseBlock-UNet~\cite{yao2020eye3dvas} & 0.8005 & - &0.8177 &0.9848 & - \\ 
DUNet~\cite{jin2019dunet} & 0.8000 & - &0.7858 &0.9880 & - \\ 
IterNet~\cite{li2020iternet} & 0.8072 & - &0.7969 & \textbf{0.9881} & -  \\ 
U-Net++~\cite{zhou2018unet++} &0.7954 &0.6606 &0.8114 &0.9847 &0.7818 \\ 
Attention U-Net~\cite{oktay2018attention} &0.7941 &0.6589 &0.8049 &0.9852 &0.7852 \\ 
\textbf{FANet} & \textbf{0.8108} & \textbf{0.6820} & \textbf{0.8544} & 0.9830 & \textbf{0.7722} \\ 
\bottomrule
\end{tabular}
\label{tab:chaseDB}
\end{table}
CHASE-DB1 is the second retinal image segmentation dataset used to evaluate our method. For this dataset, there is no official training and test split. We have used 20 images to train our model and 8 images to test as reported in the work of Li et al.~\cite{li2020iternet}. From Table~\ref{tab:chaseDB}, we can observe that our method achieved the highest F1 of 0.8108, \ac{mIoU} of 0.6820, and the highest recall of 0.8544. FANet achieved an improvement of 3.67\% in the recall compared to the \ac{SOTA} DenseBlock-UNet. 
%


\subsubsection{Results on EM dataset}
\begin{table}[!t]
\footnotesize
\caption{Results on the EM dataset~\cite{cardona2010integrated}.}
\begin{tabular}{@{}l|l|l|l|l|l@{}}
\toprule
\textbf{Method} &\textbf{F1} &\textbf{mIoU} &\textbf{Recall} &\textbf{Specificity} & \textbf{Prec.} \\ 
\midrule
U-Net~\cite{ronneberger2015u}  & - & 0.8830 & - & - & - \\ 
Wide U-Net~\cite{zhou2018unet} & - & 0.8837 & - & - & - \\ 
U-Net++~\cite{zhou2018unet++} &0.9495 &0.9038 &0.9520 &0.7875 &0.9474 \\ 
Attention U-Net~\cite{oktay2018attention} &0.9492 &0.9033 &0.9502 &0.7912 &0.9484 \\ 
\textbf{FANet}& \textbf{0.9547} & \textbf{0.9134} & \textbf{0.9568} & \textbf{0.8096} & \textbf{0.9529} \\ 
\bottomrule
\end{tabular}
\label{tab:emdataset}
\end{table}

\begin{table*}[!t]
\footnotesize
\centering
\caption{Detailed ablation study of the FANet architecture. Flop is calculated in terms of GMac. ``Rec" stands for Recall, ``Prec" stands for precision, ``Spec" stands for Specificity, ``Acc" stands for Accuracy, and ``Param" stands for total number parameters. B1 -- B4 {denote} different network configurations.~\label{tab:ablationstudy}}
\def\arraystretch{1.1}
   \begin{tabular}{@{} p{4.2cm}|p{0.7cm}|p{0.7cm}|p{0.7cm}|p{0.7cm}|p{0.7cm}|p{0.7cm}|p{0.75cm}|p{0.7cm}|p{0.7cm}|p{0.8cm}|p{1.4cm} @{}}

\toprule

\textbf{Method}  &\textbf{mIoU} & \textbf{F1}    & \textbf{Rec}& \textbf{Prec}& \textbf{Spec}& \textbf{Acc} & \textbf{F2} & \textbf{Param} & \textbf{Flops} & \textbf{FPS} & \textbf{Image Size}\\ 
\midrule

\multicolumn{12}{@{}l}{\textbf{Dataset: Kvasir-SEG}}                   \\ \midrule
Baseline (FANet without MixPool, \textbf{B1}) & 0.7732  & 0.8516 & 0.8835 & 0.8710    & 0.9783      & 0.9563   & 0.8614 & 5.76M  & 70.38 & 104.20 & $512\times 512$  \\ 
Baseline + MixPool (\textbf{B2})  & 0.6378  & 0.7302 & 0.6982 & \textbf{0.9098}   & \textbf{0.9815} & 0.9412   & 0.7039 & 7.72M  & 94.75 & 66.75 & $512\times 512$  \\ 
Baseline + MixPool(E1, D4) + feedback (\textbf{B3}) & 0.7688  & 0.8460 & 0.9047 & 0.8479    & 0.9576      & 0.9474   & 0.8699 & 5.78M  & 76.53 & 101.10 &$512\times 512$  \\ 
Baseline + MixPool + feedback (\textbf{B4}) & \textbf{0.8153}  & \textbf{0.8803} & \textbf{0.9058} & 0.9005  & 0.9794 & \textbf{0.9667} & \textbf{0.8872} & 7.72M  & 94.75 & 68.18  & $512\times 512$  \\ \midrule

\multicolumn{12}{@{}l}{\textbf{Dataset: CVC-ClinicDB}}    \\    \midrule                 
Baseline (FANet without MixPool, \textbf{B1}) & 0.8619  & 0.9166 & 0.9310 & 0.9247    & 0.9934      & 0.9877   & 0.9194 & 5.76M & 70.38 & 103.46 & $256  \times 256$  \\ 
Baseline + MixPool (\textbf{B2})                               & 0.8541  & 0.9108 & 0.9026 & 0.9296    & 0.9943      & 0.9864   & 0.9048 & 7.72M  & 94.75 & 67.490 & $256  \times 256$  \\ 
Baseline + MixPool(E1, D4) + feedback (\textbf{B3}) & 0.8729  & 0.9162 & 0.9052 & \textbf{0.9462} & 0.9941  & 0.9889 & 0.9093 & 5.78M & 76.53 & 99.03 & $256  \times 256$  \\ 
Baseline + MixPool + feedback (\textbf{B4})  & \textbf{0.8937}  & \textbf{0.9355} & \textbf{0.9339} & 0.9401  & \textbf{0.9948} & \textbf{0.9916} & \textbf{0.9342} & 7.72M  & 94.75 & 67.910  & $256  \times 256$  \\ \hline

\multicolumn{12}{@{}l}{\textbf{Dataset: 2018 Data Science Bowl}}    \\  \midrule
Baseline (FANet without MixPool, \textbf{B1}) & 0.8495  & 0.9121 & 0.9047 & 0.9283  & 0.9871  & 0.9800  & 0.9068 & 5.76M & 70.38 & 114.82 & $256  \times 256$  \\ 
Baseline + MixPool (\textbf{B2})  & 0.8158  & 0.8893 & 0.8665 & \textbf{0.9289}    & \textbf{0.9887}      & 0.9751   & 0.8733 & 7.72M  & 94.75 & 69.64  & $256  \times 256$  \\ 
Baseline + MixPool(E1, D4) + feedback (\textbf{B3}) & 0.8552  & 0.9165 & 0.9189 & 0.9199  & 0.9863  & \textbf{0.9802} & 0.9173 & 5.78M & 76.53 & 100.27 & $256  \times 256$  \\ 
Baseline + MixPool + feedback (\textbf{B4}) & \textbf{0.8569}  & \textbf{0.9176} & \textbf{0.9222} & 0.9194    & 0.9860       & 0.9800     & \textbf{0.9195} & 7.72M  & 94.75 & 69.22  & $256  \times 256$  \\ \hline

\multicolumn{12}{@{}l}{\textbf{Dataset: ISIC 2018}}             \\  \midrule
Baseline (FANet without MixPool, \textbf{B1}) & 0.7908  & 0.8647 & \textbf{0.9033} & 0.8780     & 0.9151 & 0.9151   & \textbf{0.8778} & 5.76M & 70.38 & 111.95 & $512  \times 512$  \\ 
Baseline + MixPool (\textbf{B2})  & 0.7486  & 0.8303 & 0.8049 & 0.9214    & \textbf{0.9617}      & 0.9211   & 0.8081 & 7.72M  & 94.75 & 65.91  & $512  \times 512$  \\ 
Baseline + MixPool(E1, D4) + feedback (\textbf{B3})  & \textbf{0.8078}  & \textbf{0.8780} & 0.8746 & \textbf{0.9252} & 0.9614 & \textbf{0.9374} & 0.8719 & 5.78M & 76.53 & 99.06  & $512  \times 512$  \\ 
Baseline + MixPool + feedback (\textbf{B4})  & 0.8023  & 0.8731 & 0.8650 & 0.9235    & 0.9611  & 0.9351   & 0.8630  & 7.72M  & 94.75 & 71.02  & $512  \times 512$  \\ \hline

\multicolumn{12}{@{}l}{\textbf{Dataset: DRIVE}}           \\   \midrule
Baseline (FANet without MixPool, \textbf{B1})  & 0.6912  & 0.8172 & 0.8048 & \textbf{0.8339}    & \textbf{0.9846}  & \textbf{0.9687}  & 0.8093 & 5.76M & 70.38 & 103.68 & $512  \times 512$  \\ 
Baseline + MixPool (\textbf{B2})   & 0.6895  & 0.8161 & \textbf{0.8219} & 0.8145 & 0.9820 & 0.9678   & 0.8190  & 7.72M  & 94.75 & 68.47  & $512  \times 512$  \\ 
Baseline + MixPool(E1, D4) + feedback (\textbf{B3}) & \textbf{0.6928}  & \textbf{0.8183} & 0.8124 & 0.8280 & 0.9839 & \textbf{0.9687}   & 0.8142 & 5.78M & 76.53 & 95.30  & $512  \times 512$  \\ 
Baseline + MixPool + feedback (\textbf{B4})  & 0.6927  & \textbf{0.8183} & 0.8215 & 0.8189 & 0.9826    & 0.9683   & \textbf{0.8197} & 7.72M  & 94.75 & 70.66  & $512  \times 512$  \\ \hline

\multicolumn{12}{@{}l}{\textbf{Dataset: CHASE-DB1}}                   \\   \midrule
Baseline (FANet without MixPool, \textbf{B1})                  & 0.6419  & 0.7816 & 0.7876 & 0.7768    & 0.9848      & 0.9723   & 0.7850 & 5.76M & 70.38 & 95.77  & $512  \times 512$  \\ 
Baseline + MixPool (\textbf{B2})                              & 0.5419  & 0.7009 & 0.8116 & 0.6209    & 0.9664      & 0.9565   & 0.7625 & 7.72M  & 94.75 & 65.03  & $512 \times 512$  \\ 
Baseline + MixPool(E1, D4) + feedback (\textbf{B3}) & \textbf{0.6877}  & \textbf{0.8147} & 0.8372 & \textbf{0.7948}    & \textbf{0.9855}      & 0\textbf{.9760}   & 0.8279 & 5.78M & 76.53 & 99.00  & $512 \times 512$  \\ 
Baseline + MixPool + feedback (\textbf{B4})  & 0.6820  & 0.8108 & \textbf{0.8544} & 0.7722    & 0.9830      & 0.9749   & \textbf{0.8363} & 7.72M  & 94.75 & 71.67  & $512 \times 512$  \\ \hline

\multicolumn{12}{@{}l}{\textbf{Dataset: EM}}               \\  \midrule
Baseline (FANet without MixPool, \textbf{B1}) & 0.9128  & 0.9544 & \textbf{0.9597} & 0.9495    & 0.7946      & 0.9263   & \textbf{0.9575} & 5.76M & 70.38 & 79.59  & $512 \times 512$  \\ 
Baseline + MixPool (\textbf{B2})     & 0.9121  & 0.9540  & 0.9596 & 0.9488    & 0.7918      & 0.9257   & 0.9573 & 7.72M  & 94.75 & 59.15  & $512 \times 512$  \\ 
Baseline + MixPool(E1, D4) + feedback (\textbf{B3}) & 0.9042  & 0.9497 & 0.9404 & \textbf{0.9594}    & \textbf{0.8378}      & 0.9198   & 0.9441 & 5.78M & 76.53 & 90.62  & $512 \times 512$  \\ 
Baseline + MixPool + feedback (\textbf{B4}) & \textbf{0.9134} & \textbf{0.9547} & 0.9568 & 0.9529  & 0.8096 & \textbf{0.9271}   & 0.9559 & 7.72M  & 94.75 & 70.70  & $512 \times 512$  \\ 
\bottomrule
\end{tabular}
\end{table*}

The EM dataset aims to develop an automatic \ac{ML} algorithm for the segmentation of the neural structures so that difficulties due to manual labeling can be resolved. Table~\ref{tab:emdataset} shows the quantitative results on the EM dataset. The proposed FANet also obtains F1 of 0.9547, \ac{mIoU} of 0.9134, and a recall of 0.9568. The presented results demonstrate that FANet produces \ac{SOTA} results, surpassing other recent methods in terms of \ac{mIoU} metric that was used by other methods for comparison. 
\subsection{Qualitative results}
The qualitative results on all seven datasets are presented in Figure~\ref{fig:qualitativeresult}. It can be observed that for colonoscopy datasets (Kvasir-SEG and CVC-ClinicDB), even though the initial input mask covers the entirety of the image, our model is able to prune and provide accurate masks. The same can be observed for the two retina vessel segmentation datasets, DRIVE and CHASE-DB1. It can be observed that our model is able to segment the challenging retinal vessels, including small retinal vessel bifurcations, and it well resembles the ground truth mask. For the 2018 DSB, ISIC-2018, and EM cell data, again, the input masks are finely rectified, achieving close to ground truth results by the proposed FANet model.

\subsection{Ablation study}
In this section, we ablate our model architecture and present extensive experimental results related to the effectiveness of the proposed FANet.  To evaluate the contribution of the MixPool block and the feedback, we created the following configurations:
\begin{enumerate}
    \item \textbf{Baseline} (\textbf{B1}): It refers to the FANet without the MixPool block, which means ``no feedback mechanism'' or ``iterative pruning''. We require the MixPool block to provide feedback as it unifies the attention from the network feature map and input mask (refer to Figure~\ref{fig:proposedarchitecture} (b)).
    
    \item \textbf{Baseline + MixPool} (\textbf{B2}): We integrate the MixPool block in all the encoder blocks and decoder blocks. During the inference, we directly apply the trained model weights with the Ostu thresholding (initial input mask) only once, i.e., no iterative pruning is used.
    
    \item \textbf{Baseline + MixPool(E1, D4) + Feedback} (\textbf{B3}): Here, we integrate the MixPool block in the first encoder block and the last decoder block. Feedback (iterative pruning) is used during the inference.
    
    \item \textbf{Baseline + MixPool + Feedback} (\textbf{B4}): This is the final FANet architecture, with MixPool block in all encoder and decoder blocks and the feedback (iterative pruning) mechanism is used during the inference.
    
\end{enumerate}
Table~\ref{tab:ablationstudy} presents the ablation results on these four configurations performed on all seven datasets. Below we provide detailed analyses of the use of different model architectural settings and validate them with the above described four network configurations (\textbf{B1-B4}):
\subsubsection{Effectiveness of MixPool block}
The MixPool block is an essential part of the proposed FANet architecture. It uses the previously predicted mask as the attention to improve the semantically meaningful features and allows higher-level abstractions.  The effectiveness of the MixPool block can be evaluated by comparing the network configurations B1 and B4.

From the experiments in Table~\ref{tab:ablationstudy}, we can conclude that the B4 outperforms the B1 on all the datasets. On the F1 metric, B4 shows an improvement of 2.87\% on the Kvasir-SEG dataset, 1.89\% improvement on the CVC-ClinicDB, 0.55\% improvement on the 2018 Data Science Bowl dataset, 0.84\% improvement on the ISIC 2018 dataset, 0.11\% improvement on the DRIVE dataset, 2.92\% improvement on the CHASE-DB1 dataset, and a 0.03\% improvement on the EM dataset. These performance gains are significant and thus demonstrate the effectiveness of the use of MixPool block in the proposed FANet. 


%
\subsubsection{Optimum position of MixPool block in FANet architecture}
The positioning of the MixPool is an important factor determining the performance of the model. In the FANet (B4), we integrate the MixPool block in all the encoder blocks and the decoder blocks. In B3, we integrate MixPool block in the first encoder block and the last decoder block only. To evaluate the effectiveness of the integrating MixPool block, we compare B3 with B4 in Table~\ref{tab:ablationstudy}. It can be observed that out of the seven datasets, on three datasets, i.e., Kvasir-SEG, CVC-ClinicDB, and 2018 Data Science Bowl, a  significant improvement in B4 is observed as compared to the B3. On the F1 metric, we can observe that B4 achieves an improvement of 3.43 \% on Kvasir-SEG, 1.93 \% on CVC-ClinicDB, 0.11 \% on the 2018 Data Science Bowl, and 0.5 \% on the EM dataset. 
\subsubsection{Significance of feedback during evaluation}
The proposed architecture uses the feedback information (input mask) while training. This feedback mechanism is also used during the evaluation for iterative pruning the predicted the mask. To evaluate the effectiveness of the feedback mechanism, we compare the B2 (FANet without feedback) with the B4 (FANet with feedback) in Table~\ref{tab:ablationstudy}. On all the datasets, we used feedback during inference and compared its performance with the model without feedback. We can observe that the majority of performance gains in \ac{mIoU} and F1. For Kvasir-SEG,  B4 shows a 17.75 \% improvement in the  mIoU, 15.01 \% improvement in the F1, and a 20.76 \% improvement in the Recall. Likewise, on the CVC-ClinicDB, we can see that B4 has 3.96 \% improvement in  \ac{mIoU} and 2.27 \% improvement in the F1.

\begin{figure*}[t!]
    \centering
    \includegraphics[width=\textwidth]{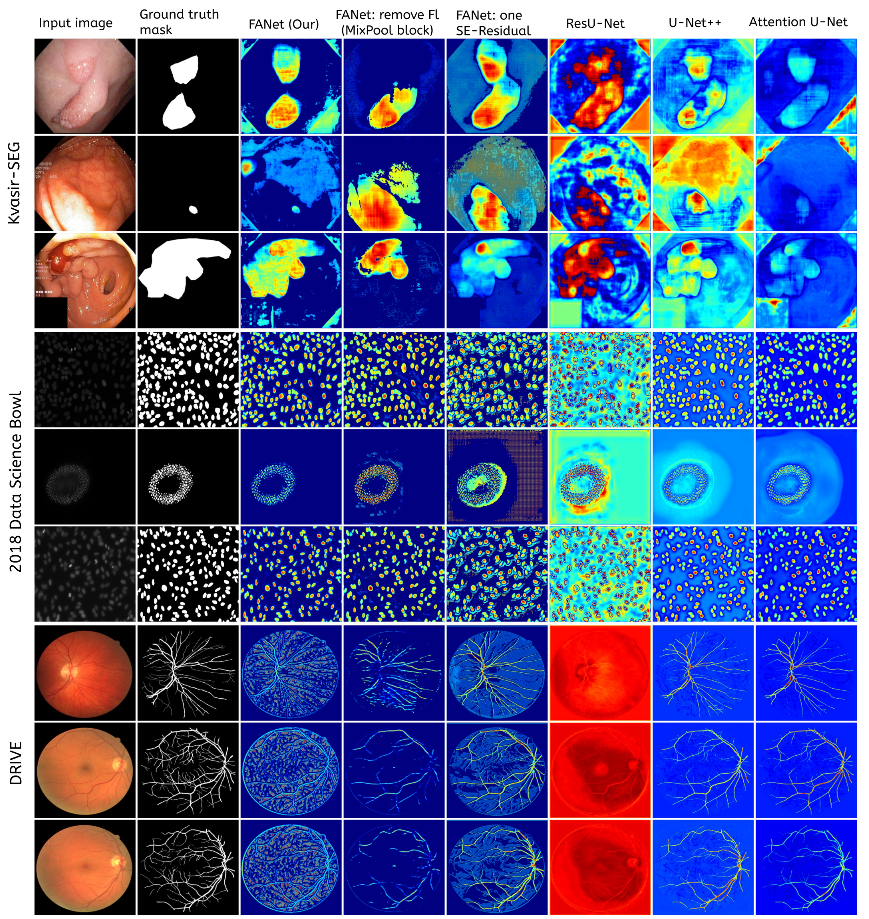}
    \caption{Comparison of the intermediate feature map of the different networks on the Kvasir-SEG, {2018 Data Science Bowl and DRIVE datasets. For each dataset, we have included three diverse images. The provided heatmaps demonstrate the impact of the weights for different networks. Here, red and yellow regions in the heat map refer to the most important features, and the blue region refers to the region of less importance. From the heat map, it can be observed that FANet has a better feature representation than other baseline networks for most of the datasets. $F_l$ represents the input feature map in the MixPool block (refer Figure~\ref{fig:proposedarchitecture}).}} 
    \label{fig:fig-heatmap-comparison}
\end{figure*}

\begin{figure}[t!]
    \centering
    \includegraphics[clip, width = \columnwidth]{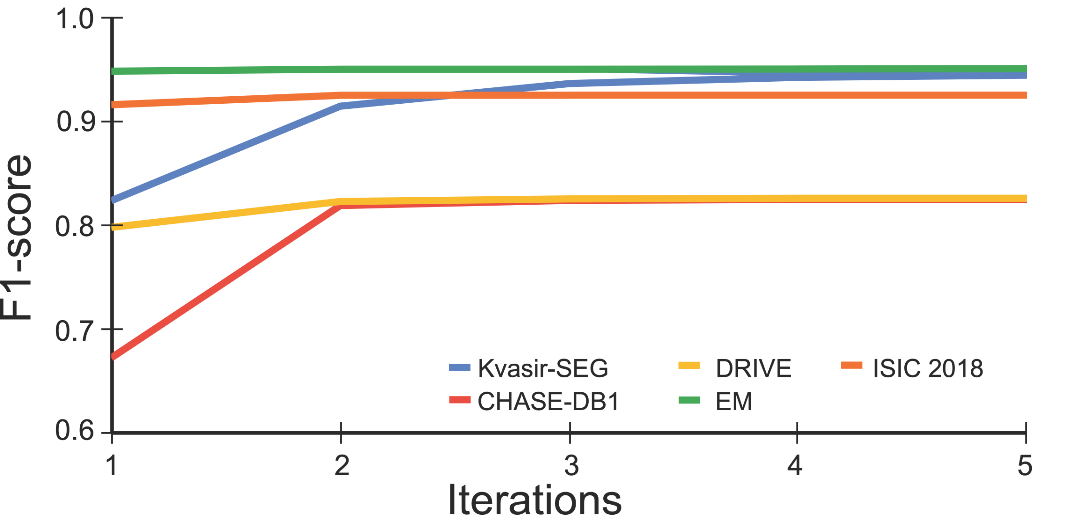}
    \caption{Iterative pruning on different dataset images.}
    \label{fig:iteration-chart}
\end{figure}

\subsection{Algorithm efficiency}
\begin{table}[t!]
\footnotesize
\caption{Algorithm complexity of our proposed FANet and other SOTA methods.}
\begin{tabular}{@{}l|l|l|l|l@{}}
\toprule
\textbf{Method} &\textbf{\shortstack{Params.\\{(million)}}} &\textbf{\shortstack{Flops\\(GMac)}} &\textbf{\shortstack{Inf. Time\\{(in ms.)}}} &\textbf{\shortstack{Image size\\{(pixels)}}} \\ 
\midrule
U-Net~\cite{ronneberger2015u} &31.04  &219.01 &3.14 &$512 \times 512$ \\ 
ResU-Net\cite{zhang2018road} &8.22  &181.68 &2.93 &$512 \times 512$  \\ 
U-Net++\cite{zhou2018unet++} &9.16 &138.6 & 4.07 &$512 \times 512$  \\ 
Attention U-Net~\cite{oktay2018attention}     &34.88  &266.54 &4.47 &$512 \times 512$  \\ 
FANet &7.72  &94.75 &8.25 &$512 \times 512$  \\ 
\bottomrule
\multicolumn{5}{l}{{Params.: parameters; Inf.: inference ; ms: milliseconds}}
\end{tabular}
\label{tab:operational-efficiency}
\end{table}

We have analyzed the algorithm efficiency in terms of the number of parameters, flops, and inference time for SOTA methods and FANet (see Table~\ref{tab:operational-efficiency}). During the architectural design, we limit the number of trainable parameters in order to minimize the computational cost of our model. The proposed FANet has only {7.72 million} parameters and 94.75 GMac flops, i.e. FANet has the least number of parameters and flops as compared to other deeper architectures. However, our inference time is higher than the other baseline networks which is due to the introduction of novel MixPool block in the FANet that incorporates additional operations such as element-wise multiplication from the readout of the RLE encoded mask that resulted in larger computational time. However, in terms of FPS per iteration this is still above 60 (see Table~\ref{tab:ablationstudy}). In the FANet, the MixPool block facilitates attention and propagation of information flow from the current learning paradigm and that of the previous epoch, which helps to achieve a performance boost (refer Table~\ref{tab:ablationstudy}). To verify the efficiency of the MixPool block, we have compared our network with and without the MixPool block in Table~\ref{tab:ablationstudy}. It is also evident that removing the MixPool block reduces the overall performance in all datasets. 

\subsection{Extended ablation study}
\begin{table}[!t]
\centering
\footnotesize
\caption{Extended ablation study demonstrating effectiveness of our proposed FANet architecture on the Kvasir-SEG dataset. Here, we ablate our network using different configuration that includes: (i) removing F$_l$ of MixPool block in {FANet (Figure~\ref{fig:proposedarchitecture}} (b)), (ii) removing and adding SE-Residual networks in FANet {(Figure~\ref{fig:proposedarchitecture} (c))}, and (iii) series concatenation of FANet in contrast to iterative mechanism.}

\begin{tabular}{@{}l|l|l|l|l@{}}
\toprule
\textbf{Method} &\textbf{F1} &\textbf{mIoU} &\textbf{Recall} &\textbf{Prec.} \\ 
\midrule
MixPool block (w/o F$_l$) &0.6527 &0.5523 &0.7175 &0.7255  \\ 
One SE block only &0.7641 &0.6810 &\textbf{0.9531} &0.7133  \\ 
Three SE blocks &0.8570 &0.7908 &0.8974 &0.8768  \\ 

Four SE blocks &0.8158 &0.7417 &0.8528 &0.8605  \\
Four FANet (in series) &0.8682 &0.8004 &0.8994 &0.8913  \\
\textbf{Two SE in FANet (Ours)} &\textbf{0.8803} &\textbf{0.8153} &0.9058 &\textbf{0.9005} \\
\bottomrule
\end{tabular}
\label{tab:extended-ablation-study}
\end{table}
We have performed an extended ablation study to demonstrate the architectural effectiveness of the proposed FANet. Here, we begin with the experimental verification of the MixPool block by removing its certain components. From the Table~\ref{tab:extended-ablation-study}, we can observe a performance drop, F1 drops by $22.76$\% and mIoU drops by $26.3$\% when feature map $F_l$ is not used during feature concatenation. In order to justify the use of two SE-Residual block in our proposed FANet, we conducted studies by removing and adding SE-Residual blocks from the FANet. We first began by removing a SE-Residual block for which we observed an $11.62$\% drop in F1 and $13.43$\% drop in mIoU. 

Further, we modified the FANet architecture by adding three SE-Residual blocks and we observed again a decrease in the performance. For this case, F1 drops by $2.33$\% and mIoU by $2.45$\%. Next, we added one more SE-Residual block and a severe performance drop can be observed. The F1 dropped by $6.45$\% and the mIoU drops by $7.36$\%. In our proposed architecture we used an iterative pruning. However, we experimented an alternative strategy by concatenating the four FANet together in a series. From this experiment we observed a drop of F1 by $1.21$\% and $1.49$\% drop in mIoU  with nearly four times increase in the number of trainable parameters.
%
\section{Discussion}
While deep learning semantic segmentation has been widely implemented, to the best of our knowledge, only direct inference strategies have been published till date. In this work, we utilize a segmentation map pruning mechanism that demonstrates a clear advantage over the current \ac{SOTA} models due to its ability to self-rectify the predicted mask during the evaluation process (see Table~\ref{tab:kvasirseg}-Table~\ref{tab:emdataset}). The process of self-rectification or iterative pruning helps to improve the performance of the proposed FANet architecture. This improvement is due to the feedback provided by the input mask in the MixPool block which is further validated from our two ablation studies (Table~\ref{tab:ablationstudy} and Table~\ref{tab:extended-ablation-study}). Furthermore, a joint configuration together with mask and the feature embeddings allow learning to achieve better feature representation of target regions and learning to adjust weights dependent on the input mask. This establishes an effective pruning mechanism of the network enabling input mask to be steered in the direction of the relevant {learned} features of the network. Additionally, it can capture the variability in datasets (e.g., shape distributions, surface morphology etc.), allowing the network to rectify the predicted/input masks.

Table~\ref{tab:ablationstudy} shows the complete ablation study of the MixPool block in the FANet architecture. In this ablation, we provide experimental results with (proposed network, B4) and without the MixPool block (B1). Here, B1 refers to the ``no feedback mechanism'' as no MixPool block is applied. In the proposed FANet, we require the MixPool block to provide feedback through a unified attention mechanism taking into account the network feature map and the input mask from the previous epoch (refer to Figure~\ref{fig:proposedarchitecture} (b)). However, for the MixPool block without feedback (i.e., B2), we provide the attention from the generated feature map and the input mask but we do not perform the iterative pruning during the evaluation. Thus, even though B2 and B4 networks have the same number of parameters {(7.72 Million parameters)}, the removal of the feedback mechanism affects the algorithm performance (see Table~\ref{tab:ablationstudy}). SE-Residual blocks that serve as a self-attention mechanisms on feature channels by performing global average pooling followed by multi layer perceptron that allows to explicitly model the interdependencies between feature channels. Further, in our network we introduce spatial attention mechanisms. Multiple SE-residual blocks allow to learn complex non-linear feature interdependencies (also see Table~\ref{tab:extended-ablation-study} for different combinations of SE-Residual blocks).
Further, other ablation experiments such as series concatenation of FANet and removal of $F_l$ layer in MixPool block in Table~\ref{tab:extended-ablation-study} showed that the proposed FANet achieves the highest performance. This justifies the importance of different components integrated in the proposed FANet architecture. {Further, qualitative results in Figure~\ref{fig:fig-heatmap-comparison} demonstrate the effectiveness of our network over different configurations, for example, removing of $F_l$ layer in mixpool block and using only one SE-Residual block. Also, it can be observed that FANet has more apparent segmentation maps, that is easily distinguishable regions from the background, than the SOTA methods.} 

With the introduction of the iterative pruning in our FANet architecture, we introduce a new hyperparameter, i.e., the number of iterations during the evaluation. The optimal number of iterations is 10, which was empirically established across datasets. The computed number of iterations is the same for all datasets. For this, we have plotted a graph (Figure~\ref{fig:iteration-chart}) showing the iterative pruning on different dataset images. From the graph, it is observed that there is a significant improvement from iteration 1 to 5. However, from 5 to 10 iterations, there is a minor to negligible improvement. Thus, we considered the highest of 10 iterations during the evaluation. The iterative pruning over the input image increases the inference time. However, this process allows us to refine the predicted segmentation masks, unlike most current methods. For obtaining a better trade-off between efficiency and accuracy, we advise using a lesser number of iterations. We plot the F1 score for different dataset images for five iterations during evaluation. Figure~\ref{fig:iteration-chart} shows that our proposed FANet benefits with just two iterations. Additionally, we have used the NVIDIA GTX 1050 Ti (released in 2016) for inference, and thus using a more recent GPU with higher performance can provide better inference time.
\section{\color{black}{Conclusion}}
\label{conclusion}
\color{black}{With the FANet architecture, we proposed a novel approach for biomedical image segmentation that can self-rectify the predicted masks. By introducing a feedback mechanism, we achieved an improvement on seven publicly available biomedical datasets when compared with existing \ac{SOTA} methods. Our approach requires far fewer epochs for training and is well-suited to diverse biomedical imaging datasets. The feedback mechanism integrated in the FANet design effectively acts as hard attention that is used with the existing feature maps to boost the strength of feature representations. The experimental results demonstrate that the proposed architecture achieves accurate and consistent segmentation results across several biomedical imaging datasets despite its simple and straightforward network architecture. The ablation study also reveals that FANet requires less training time to achieve near \ac{SOTA} performance. In the future, we will use a contrastive learning approach to improve the performance of FANet further and test it on additional multimodal biomedical images.}

\section*{Acknowledgment}
D. Jha is funded by the Research Council of Norway project number 263248 (PRIVATON).  S. Ali is supported by the National Institute for Health Research (NIHR) Oxford Biomedical Research Centre (BRC). The computations in this paper are performed on equipment provided by the Experimental Infrastructure for Exploration of Exascale Computing (eX3), which is financially supported by the Research Council of Norway under contract 270053. The views expressed are those of the author(s) and not necessarily those of the NHS, the NIHR, or the Department of Health. 

\bibliographystyle{IEEEtran}
\bibliography{references} 

\end{document}